\documentclass[sigconf]{acmart}

\usepackage{subcaption}
\usepackage{threeparttable}
\copyrightyear{2025} 
\acmYear{2025} 
\setcopyright{cc} 
\setcctype{by} 
\acmConference[MM '25]{Proceedings of the 33rd ACM International Conference on Multimedia}{October 27--31, 2025}{Dublin, Ireland} 
\acmBooktitle{Proceedings of the 33rd ACM International Conference on Multimedia (MM '25), October 27--31, 2025, Dublin, Ireland}
\acmDOI{10.1145/3746027.3758235} 
\acmISBN{979-8-4007-2035-2/2025/10}

\begin{document}

\title{RSVLM-QA: A Benchmark Dataset for Remote Sensing Vision Language Model-based Question Answering}


\author{Xing Zi}
\email{xing.zi-1@uts.edu.au}
\orcid{0009-0001-4265-2205}
\affiliation{%
  \institution{School of Computer Science, University of Technology Sydney}
  \city{Sydney}
  \state{New South Wales}
  \country{Australia}}

\author{Jinghao Xiao}
\email{jinghao.xiao@student.uts.edu.au}
\orcid{0009-0000-0837-9068}
\affiliation{%
  \institution{School of Computer Science, University of Technology Sydney}
  \city{Sydney}
  \state{New South Wales}
  \country{Australia}}

\author{Yunxiao Shi}
\email{yunxiao.shi@student.uts.edu.au}
\orcid{0000-0002-1516-015X}
\affiliation{%
  \institution{SEDE, University of Technology Sydney}
  \city{Sydney}
  \state{New South Wales}
  \country{Australia}}

\author{Xian Tao}
\authornote{Corresponding Author Xian Tao. Xian Tao is with the Institute of Automation, Chinese Academy of Sciences, Beijing, China. Email: taoxian2013@ia.ac.cn}
\email{taoxian2013@ia.ac.cn}
\orcid{0000-0001-5834-5181}
\affiliation{%
  \institution{Institute of Automation, Chinese Academy of Sciences}
  \city{Beijing}
  \country{China}}

\author{Jun Li}
\email{jun.li@uts.edu.au}
\orcid{0000-0002-1336-2241}
\affiliation{%
  \institution{School of Computer Science, University of Technology Sydney}
  \city{Sydney}
  \state{New South Wales}
  \country{Australia}}

\author{Ali Braytee}
\email{ali.braytee@uts.edu.au}
\orcid{0000-0003-2561-6496}
\affiliation{%
  \institution{School of Computer Science, University of Technology Sydney}
  \city{Sydney}
  \state{New South Wales}
  \country{Australia}}

\author{Mukesh Prasad}
\email{mukesh.prasad@uts.edu.au}
\orcid{0000-0002-7745-9667}
\affiliation{%
  \institution{School of Computer Science, University of Technology Sydney}
  \city{Sydney}
  \state{New South Wales}
  \country{Australia}}

\renewcommand{\shortauthors}{Zi et al.}

\begin{abstract}
Visual Question Answering (VQA) in remote sensing (RS) is pivotal for interpreting Earth observation data. However, existing RS VQA datasets are constrained by limitations in annotation richness, question diversity, and the assessment of specific reasoning capabilities. This paper introduces \textbf{R}emote \textbf{S}ensing \textbf{V}ision \textbf{L}anguage \textbf{M}odel \textbf{Q}uestion \textbf{A}nswering (RSVLM-QA) dataset, a new large-scale, content-rich VQA dataset for the RS domain. RSVLM-QA is constructed by integrating data from several prominent RS segmentation and detection datasets: WHU, LoveDA, INRIA, and iSAID. We employ an innovative dual-track annotation generation pipeline. Firstly, we leverage Large Language Models (LLMs), specifically GPT-4.1, with meticulously designed prompts to automatically generate a suite of detailed annotations including image captions, spatial relations, and semantic tags, alongside complex caption-based VQA pairs. Secondly, to address the challenging task of object counting in RS imagery, we have developed a specialized automated process that extracts object counts directly from the original segmentation data; GPT-4.1 then formulates natural language answers from these counts, which are paired with preset question templates to create counting QA pairs. RSVLM-QA comprises 13,820 images and 162,373 VQA pairs, featuring extensive annotations and diverse question types. We provide a detailed statistical analysis of the dataset and a comparison with existing RS VQA benchmarks, highlighting the superior depth and breadth of RSVLM-QA's annotations. Furthermore, we conduct benchmark experiments on Six mainstream Vision Language Models (VLMs), demonstrating that RSVLM-QA effectively evaluates and challenges the understanding and reasoning abilities of current VLMs in the RS domain. We believe RSVLM-QA will serve as a pivotal resource for the RS VQA and VLM research communities, poised to catalyze advancements in the field. The dataset, generation code, and benchmark models are publicly available at \url{https://github.com/StarZi0213/RSVLM-QA}.
\end{abstract}


\received{30 May 2025}
\received[accepted]{31 July 2025}

\maketitle

\section{Introduction}
\label{sec:introduction}

Recent years have witnessed remarkable advancements in Vision Language Models (VLMs), fundamentally transforming their capacity to integrate visual data with nuanced natural language understanding~\cite{VLM_survey1}. Within the critical domain of Earth observation, Remote Sensing Visual Question Answering (RSVQA) has emerged as a pivotal application that enables models to interpret complex remote sensing imagery and respond to intricate textual inquiries. This capability is of paramount importance for diverse real-world applications, including environmental surveillance, urban planning, and disaster response~\cite{VRSBench, RSVQA-HR, RSICap, CRSVQA}.

Initial explorations in RSVQA laid crucial foundations, typically by sourcing imagery from public remote sensing archives and adapting contemporary vision-language architectures (e.g., CNN encoders with RNN decoders for image description generation~\cite{UCM_Caption, chatearthnet, LRS-VQA}. The corresponding question-answer pairs were often compiled through laborious manual annotation, restrictive template-based scripting, or elementary rule-based systems drawing from extant geospatial databases\cite{RSICD, TAMMI, LRS-VQA}. These pioneering efforts were crucial to the field's inception. However, they also revealed challenges, such as the difficulty of building scalable datasets, deeply understanding complex remote sensing images, and developing diverse, sophisticated reasoning tasks.

\begin{figure*}[t!]
    \centering
    \includegraphics[width=1\textwidth]{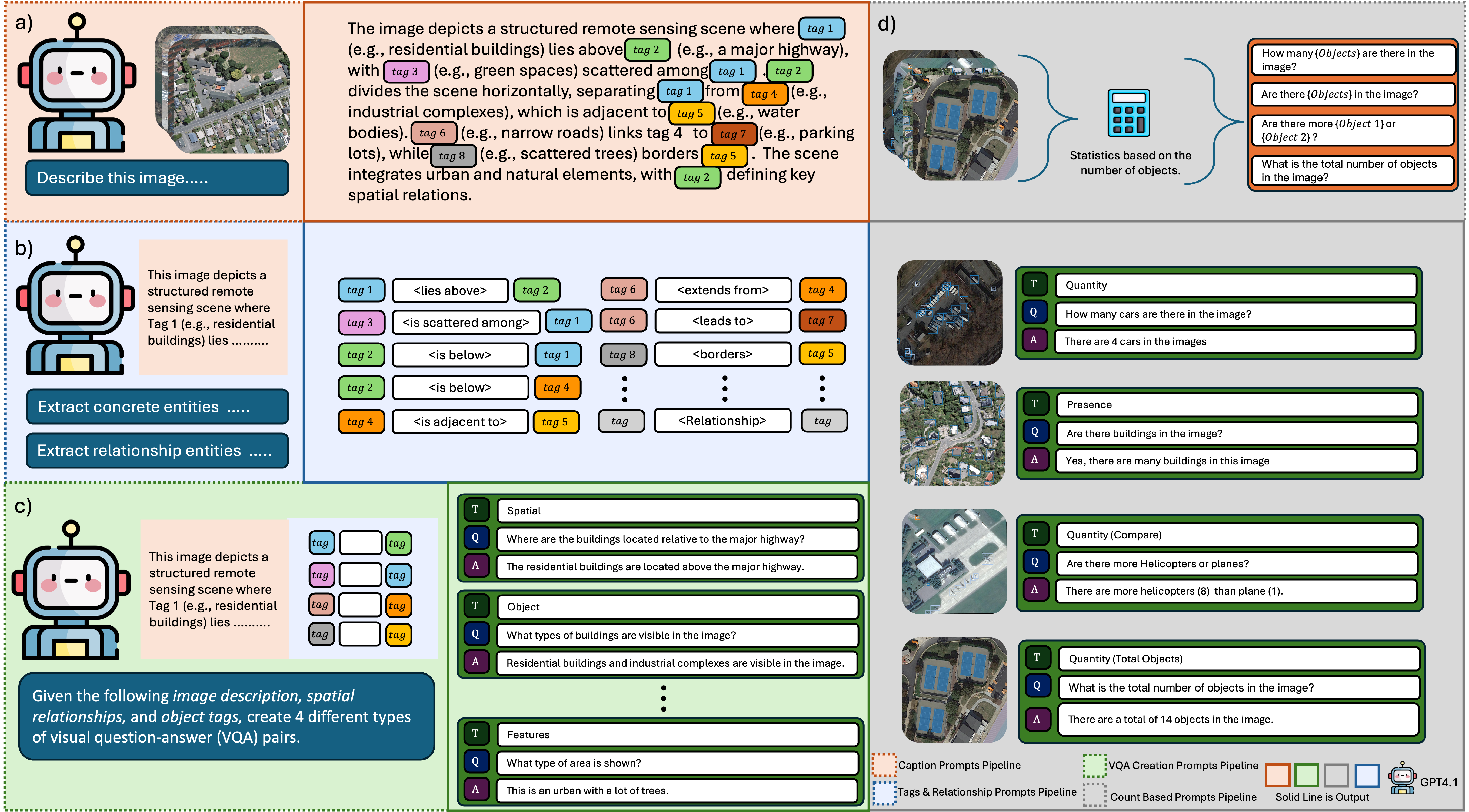}
    \caption{Illustration of the multi-stage, Large Language Model-driven (GPT-4.1) annotation and Visual Question Answering (VQA) generation workflow for the RSVLM-QA dataset. The process first, in stage (a), utilizes a Large Language Model (GPT-4.1) to process input remote sensing images, generating rich and detailed textual descriptions (Captions). Next, in stage (b), concrete entities (objects) and their spatial relationships are extracted from the textual descriptions generated in stage (a). Then, the information generated in stages (a) and (b) is jointly fed into stage (c) to construct diverse, description-based VQA pairs aimed at addressing spatial reasoning, object attributes, and feature analysis. Concurrently, in stage (d), a count-based prompts pipeline generates precise, quantitative VQA pairs related to each object's enumeration (e.g., object counts, existence checks, quantity comparisons, and total object queries) by leveraging statistics derived from object information within four semantic segmentation or object detection source datasets. Finally, all generated VQA pairs are consolidated into a single, unified dataset.}
    \label{fig:llm_annotation_pipeline}
\end{figure*}

These foundational limitations give rise to several pressing challenges impeding the ongoing advancement of the RSVQA field. Firstly, the prohibitive expense and extensive time commitment required to curate large-scale RSVQA datasets replete with detailed and multifaceted annotations remain a primary bottleneck \cite{RS5M, remoteclip}. Secondly, many existing datasets suffer from limited annotation richness, often lacking. For instance, comprehensive image narratives, structured spatial relationship data, or fine-grained semantic tagging and exhibit a narrow diversity in question typology, especially for inquiries demanding nuanced comprehension beyond rudimentary object presence detection. \cite{challenge1,challenge2,challenge3}. Thirdly, there is a critical underrepresentation of tasks that rigorously assess models' fine-grained understanding and quantitative reasoning faculties as applied to remote sensing imagery, such as precise object enumeration or the interpretation of complex spatial interdependencies. Exacerbating these issues, remote sensing images inherently present unique interpretive difficulties due to their vast spatial scales, high resolutions, the prevalence of diminutive objects, and intricate land cover taxonomies, all of which diverge significantly from natural image characteristics and necessitate a more sophisticated level of fine-grained understanding from VLMs~\cite{VRSBench, RS5M}. For a detailed comparison of our dataset with the current RS dataset, and check Comparison of Remote Sensing VQA and Image Captioning Datasets section.

To address these challenges and advance the RSVQA field, this paper introduces RSVLM-QA, a large-scale, multi-source, and richly annotated VQA dataset for remote sensing. RSVLM-QA is constructed by integrating four diverse remote sensing datasets---the WHU Building Dataset~\cite{WHU}, LoveDA~\cite{LoveDA}, the INRIA Aerial Image Labeling Dataset~\cite{AerialDataset}, and iSAID~\cite{isaid}. And employing a novel LLM-driven (GPT-4.1) pipeline. This pipeline first generates detailed textual annotations (captions, spatial relations, and semantic tags) using meticulously designed prompts. Subsequently, it utilizes a dual-track mechanism for VQA pair generation: one track creates complex, description-based questions from the LLM-generated annotations. While the other performs automated, ground-truth based generation of four types of precise object counting questions (direct count, existence, comparison, and total object queries) directly from source labels, ensuring quantitative accuracy.

RSVLM-QA comprises 13,820 images and 162,373 VQA pairs, featuring diverse question types (6 broad categories) and extensive textual annotations. We benchmarked RSVLM-QA on six leading VLMs, demonstrating its effectiveness and challenging nature. The dataset, generation code, and benchmark results are publicly available to foster further research.

The primary contributions of this work are:
\begin{enumerate}
    \item \textbf{Automated LLM-Driven Dataset Construction:} Utilizing GPT-4.1 with sophisticated prompt engineering, we develop a scalable pipeline to generate detailed annotations (captions, spatial relations, tags) and diverse VQA pairs, with rigorous human validation ensuring data coherence and accuracy.
    \item \textbf{Dual-Mode VQA for Reasoning and Precision:} Our innovative approach merges LLM-generated descriptive questions to foster complex reasoning with ground-truth-based counting questions for precise quantitative remote sensing analysis.
    \item \textbf{Multi-Faceted Annotations for Broad Research:} RSVLM-QA provides rich captions, structured relations, and semantic tags, enabling diverse multi-modal tasks such as image captioning and scene understanding.
    \item \textbf{Comprehensive Benchmark Suite:} We establish six benchmarks, including five VQA categories (object recognition, feature analysis, spatial reasoning, quantitative queries, presence verification) and detailed image captioning, to evaluate advanced visual language models.
\end{enumerate}

\section{Methodology: Principled Construction of the RSVLM-QA Dataset}
\label{sec:methodology}

The RSVLM-QA dataset was developed using a principled methodology designed to address key limitations in existing RSVQA datasets, such as semantic richness, question diversity, and visual grounding fidelity. Our approach synergistically integrates the advanced generative capabilities of Large Language Models (LLMs), notably GPT-4.1, with meticulous data curation and stringent quality assurance protocols. The following discussion elaborates on the core principles and strategic design choices underpinning this pipeline, complementing the procedural illustration in Figure~\ref{fig:llm_annotation_pipeline}.

\subsection{Source Datasets for Visual Diversity}
\label{ssec:source_datasets}

The visual foundation of RSVLM-QA is established by strategically integrating four distinct public remote sensing datasets—the WHU Building Dataset, LoveDA, the INRIA Aerial Image Labeling Dataset, and iSAID. To maximize VQA diversity across object recognition (\textbf{Objects}), feature analysis (\textbf{Features}), spatial reasoning (\textbf{Spatial}), quantitative queries (\textbf{Quantity}), and existence verification (\textbf{Presence}). Each dataset brings unique strengths to these aspects. The WHU Building Dataset offers numerous building instances, primarily strengthening quantitative tasks (e.g., building counts) and presence verification for this class, while its varied scenes contribute to building feature diversity~\cite{WHU}. LoveDA enriches RSVLM-QA with distinct urban and rural features and diverse land covers (such as forests, water bodies, and agricultural areas alongside built-up regions); this enables nuanced spatial reasoning about area arrangements and provides varied object types~\cite{LoveDA}. The INRIA Dataset, with its focus on multiple international cities, supplies diverse urban features and building objects, thereby supporting spatial analysis in dense cityscapes and assessments of building quantities~\cite{AerialDataset}. Finally, iSAID, with its 15 distinct categories and over 655,000 instances (many of which are small objects), crucially expands object diversity. This richness enables complex quantitative tasks for diverse items, robust detection of various object types, and intricate spatial analysis in feature-rich, dense scenes.~\cite{isaid}

By integrating these datasets, RSVLM-QA encompasses a diverse set of images spanning broad geographical regions, varied scene types (e.g., dense urban cores, sparse rural areas, industrial zones, natural landscapes), and numerous object categories (e.g., buildings, roads, vehicles, vessels, aircraft, sports grounds, water bodies, vegetation). This diversity is essential for training and evaluating VLMs to generalize across diverse remote sensing contexts. All prompt templates are available at \url{https://github.com/StarZi0213/RSVLM-QA/blob/main/Prompts.md}.

\subsection{LLM-Powered Rich Semantic Annotation}
\label{ssec:llm_semantic_annotation}

A key part of RSVLM-QA is creating detailed descriptions from diverse source images, as outlined in Section~\ref{ssec:source_datasets}. This goes beyond simple object labels to deeply understand scenes. It relies on two main steps powered by LLMs, shown as stages (a) and (b) in Figure~\ref{fig:llm_annotation_pipeline}:

We first leverage GPT-4.1 to generate detailed, narrative captions for each image. Advanced prompt engineering directs GPT-4.1 to identify salient objects, their spatial configurations, key landscape features, and critical directional cues. This strategy requires qualitative descriptors, such as 'a significant portion,' instead of precise percentages, to ensure natural language. It also enforces a 'Total-subtotal' caption structure for consistent depth. A core principle is emphasizing visually verifiable content to minimize model hallucination.

In the second phase, we convert the rich narrative captions into a structured format suitable for downstream processing. This involves an information extraction step, where we parse the LLM-generated captions to identify key object entities and define their explicit spatial interrelationships. The resulting structured output—entities and their relations—forms a critical layer of machine-interpretable semantic knowledge about the scene.

\begin{figure}[t!]
  \centering
  \begin{subfigure}[b]{0.45\textwidth}
    \centering
    \includegraphics[width=\linewidth]{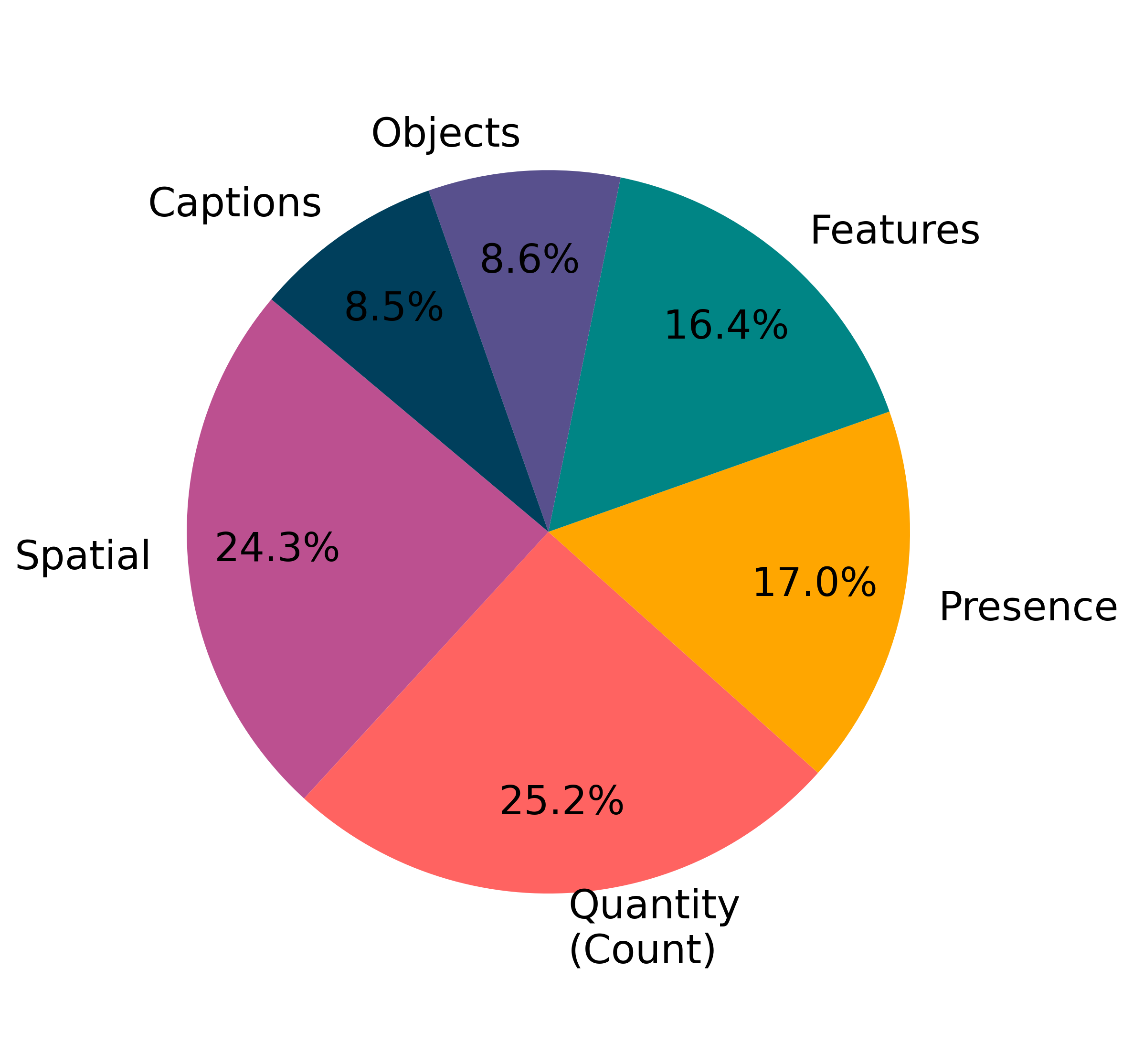} 
    \subcaption{Proportion of VQA pairs by question type.}
    \label{fig:qtype_dist_pie}
  \end{subfigure}
  \hfill
  \begin{subfigure}[b]{0.45\textwidth}
    \centering
    \includegraphics[width=\linewidth]{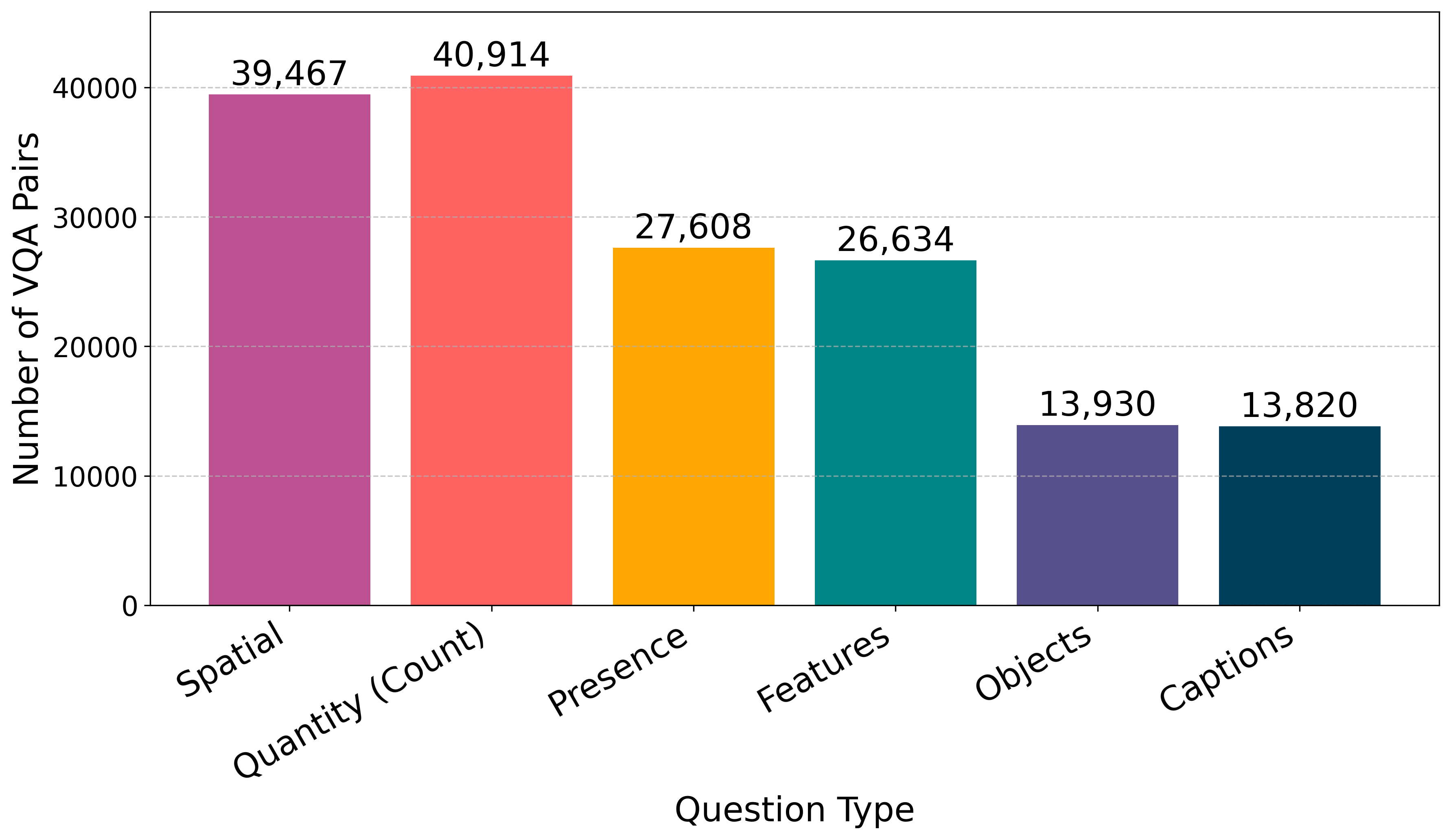} 
    \subcaption{Number of VQA pairs by question type.}
    \label{fig:qtype_dist_bar}
  \end{subfigure}
  \caption{Distribution of VQA pairs across the 6 main question types in RSVLM-QA, illustrating both the proportional share (a) and absolute counts (b) for each category. \textit{Quantity (Count)} and \textit{Spatial} questions are the most prevalent.}
  \label{fig:qtype_dist_combined}
\end{figure}
\subsection{Dual-Track VQA Generation for Semantic Depth and Quantitative Precision}
\label{ssec:dual_track_vqa}

To ensure RSVLM-QA supports a broad spectrum of reasoning capabilities, we implemented a dual-track VQA generation strategy, as shown in stages (c) and (d) of Figure~\ref{fig:llm_annotation_pipeline}. This approach targets two goals: achieving semantic depth in qualitative understanding and ensuring high precision in quantitative assessments:

The first track focuses on Description-Driven VQA. It leverages the rich captions and structured entity-relationship data generated in the preceding annotation phases (stages a and b) to automatically formulate complex questions. These VQA pairs are engineered to probe sophisticated understanding, requiring nuanced spatial reasoning, fine-grained attribute identification, and holistic scene feature analysis.

The second, parallel track ensures Ground-Truth Anchored Quantitative VQA. Recognizing the critical need for factual accuracy in numerical queries, this track directly utilizes ground-truth object information and statistics derived from the annotations available in the four source semantic segmentation or object detection datasets. An automated workflow systematically generates precise VQA pairs targeting various forms of object enumeration, including direct counts, existence verification, comparative quantification, and aggregate tallies.

\subsection{Rigorous Data Curation and Quality Assurance}
\label{ssec:curation_quality_assurance}
Ensuring the integrity and utility of RSVLM-QA is paramount. Thus, all generated captions and VQA pairs undergo a stringent two-stage quality assurance protocol. Initially, GPT-4.1 serves as an intelligent screening agent, following specific instructions to flag VQA pairs with potential ambiguities, factual inconsistencies with remote sensing norms, nonsensical content, or ill-posed questions (e.g., overly subjective queries or those requiring imperceptible details). Then, trained human experts perform meticulous manual verification and adjudication on flagged items and a substantial random sample of unflagged data. This iterative refinement process filtered approximately 14\% of the initial VQA pairs. It enhances coherence, contextual relevance, and factual accuracy, establishing RSVLM-QA as a robust and fair benchmark for advanced VLM evaluation in Earth observation.

\subsection{RSVLM-QA: Dataset Characteristics and Statistical Analysis}
\label{ssec:dataset_statistics}

The RSVLM-QA dataset is built by integrating and enhancing multiple remote sensing datasets, as outlined in Section~\ref{ssec:source_datasets}. It features a large scale, rich annotations, and a diverse range of question types. This section provides a detailed statistical analysis to highlight these qualities, demonstrating its value for robust VLM evaluation in Earth observation.

\subsubsection{Overall Statistics}
RSVLM-QA contains 13,820 images and 162,373 Visual Question Answering (VQA) pairs, organized into 6 main question types. Its textual components include a vocabulary of approximately 5,700 unique words. Table~\ref{tab:overall_stats} offers a comprehensive overview, illustrating the depth of both textual and structured annotations.

\begin{table}[htbp]
  \caption{Overall Statistics of the RSVLM-QA Dataset.}
  \label{tab:overall_stats}
  \small 
  \begin{tabular}{lr}
    \toprule
    \textbf{Metric} & \textbf{Value} \\
    \midrule
    Total Images & 13,820 \\
    Total VQA Pairs & 162,373 \\
    Question Types & 6 \\
    Vocabulary Size (Unique Words) & $\sim$5,700 \\ 
    \midrule
    Avg. Relations per Image & 5.63 \\
    Avg. Tags (Entities) per Image & 10.62 \\
    \midrule
    Avg. Question Length (words) & 9.23 \\
    Avg. Answer Length (words) & 18.80 \\
    \midrule
    Total Caption Sentences & 62,539 \\
    Avg. Sentences per Caption & 4.67 \\
    Avg. Caption Length (words) & 124.25 \\ 
    \bottomrule
  \end{tabular}
\end{table}

\subsubsection{Richness of Structured Annotations}
Beyond VQA pairs and captions, RSVLM-QA provides dense structured annotations in the form of object tags (entities) and spatial relations. Table~\ref{tab:structured_annotations} details the statistics for these annotations, underscoring the dataset's capacity to support tasks requiring fine-grained scene graph understanding. On average, each image is annotated with 5.63 spatial relations and 10.62 object tags.

\begin{table}[htbp]
  \caption{Statistics of Structured Annotations (Relations and Tags) in RSVLM-QA.}
  \label{tab:structured_annotations}
  \small
  \begin{tabular}{lrrrrr}
    \toprule
    \textbf{Type} & \textbf{Total} & \textbf{Avg. per Image} & \textbf{Median} & \textbf{Max} & \textbf{Min} \\
    \midrule
    Relations & 77,823 & 5.63 & 6 & 19 & 1 \\
    Tags (Entities) & 146,814 & 10.62 & 11 & 29 & 1 \\
    \bottomrule
  \end{tabular}
\end{table}

\begin{table}[htbp]
  \centering
  \begin{threeparttable}
    \caption{Distribution and Characteristics of Question Types in RSVLM-QA. Statistics refer to average question (Avg. Q) and answer (Avg. A) lengths in words.}
    \label{tab:vqa_type_distribution}
    \footnotesize 
    \begin{tabular}{lrrr}
      \toprule
      \textbf{Category} & \textbf{Total VQA Pairs} & \textbf{Avg. Q Length} & \textbf{Avg. A Length} \\
      \midrule
      Spatial & 39,467 & 10.58 & 12.35 \\
      Quantity & 40,914 & 9.25 & 10.45 \\
      Presence & 27,608 & 7.17 & 5.17 \\
      Features & 26,634 & 9.95 & 13.58 \\
      Objects & 13,930 & 10.30 & 10.81 \\
      Captions & 13,820 & 7.00 & 107.82 \\
      \midrule
      \textbf{Total} & \textbf{162,373} & \textbf{9.23} & \textbf{10.58}\tnote{a} \\ 
      \bottomrule
    \end{tabular}
    \begin{tablenotes}
      \item[a] \footnotesize The 'Total' for 'Avg. A Length' excludes the 'Captions' category due to its significantly different answer length characteristic, providing a more representative average for other VQA types.
    \end{tablenotes}
  \end{threeparttable}
\end{table}

\subsubsection{Diversity of Question Types}
A defining feature of RSVLM-QA is its diverse set of question types, designed to evaluate varied reasoning capabilities of VLMs. The dataset is divided into six broad categories. Their distribution and characteristics are outlined in Table~\ref{tab:vqa_type_distribution}. Figure~\ref{fig:qtype_dist_combined} illustrates this distribution, showing both proportional representation and absolute counts for each type. Notably, \textit{Quantity (Count)} and \textit{Spatial} questions dominate, emphasizing the dataset’s focus on essential remote sensing reasoning skills. The 'Captions' category, where questions elicit detailed descriptions, features answers with a significantly higher average length, reflecting their narrative nature.

\subsubsection{Textual Complexity}
The textual components of RSVLM-QA, including questions, answers, and captions, exhibit notable complexity. As shown in Table~\ref{tab:overall_stats}, the average question length is 9.23 words, and the average answer(w/o caption) length is 10.58 words, indicating that the VQA pairs often require more than trivial responses. The detailed captions are substantially longer, averaging 124.25 words and approximately 4.67 sentences per caption. This linguistic richness, combined with a vocabulary of around 5,700 unique words, provides a challenging testbed for nuanced language understanding in VLMs.

In summary, the statistical profile of RSVLM-QA underscores its capacity as a large-scale, richly annotated, and diverse benchmark. It is well-suited to drive future research in developing VLMs with sophisticated reasoning and understanding capabilities for complex remote sensing imagery.

\section{Experiments}
\label{sec:experiments}

To validate the utility and challenging nature of RSVLM-QA, we conduct comprehensive benchmarking experiments using several state-of-the-art Vision Language Models (VLMs). This section details the experimental setup, presents the performance results, and provides an analysis of the findings, underscoring the capabilities and current limitations of VLMs in the remote sensing domain.

\subsection{Experimental Setup}
\label{ssec:experimental_setup}

\subsubsection{Benchmarked VLMs.} We selected six prominent VLMs for evaluation: Gemma3~\cite{gemma3}, InternVL3~\cite{internvl3}, LLaVA~\cite{llava},BLIP-2~\cite{blip-2}, Qwen2.5-VL~\cite{qwen2-vl}, and Ovis2~\cite{ovis2}. These models represent a diverse range of contemporary VLM architectures and were chosen for their strong performance on general vision-language benchmarks and their public availability. All VLMs were evaluated in a zero-shot inference mode on RSVLM-QA. This approach rigorously assesses their intrinsic generalization capabilities to our specific remote sensing VQA and captioning tasks without any task-specific fine-tuning on our dataset, thereby providing a stringent test of their adaptability to the Earth observation domain.
\begin{figure*}[t!]
  \centering
  \includegraphics[width=0.95\textwidth]{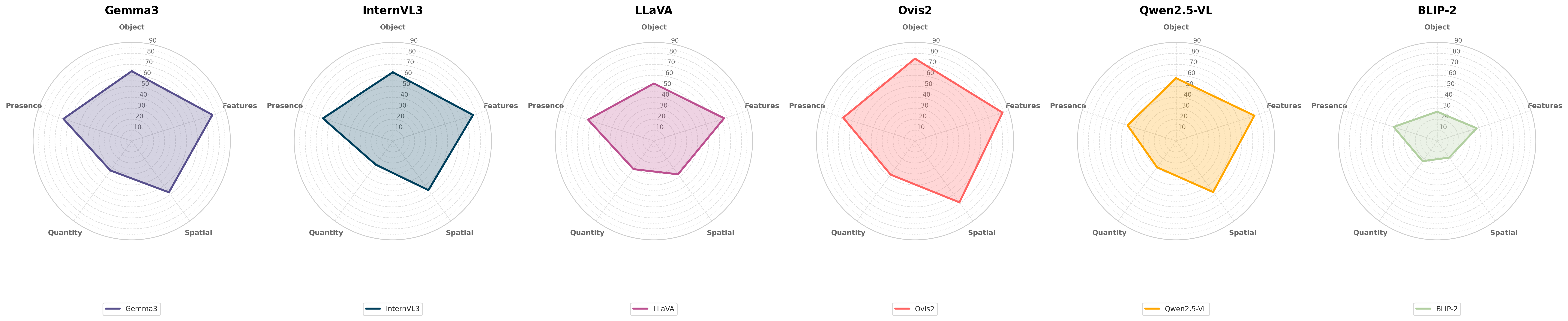} 
  \caption{Zero-shot VQA performance of six leading VLMs on five core categories of RSVLM-QA: Object Recognition, Feature Analysis, Spatial Reasoning, Quantity, and Presence, evaluated by GPT-4.1. Scores represent VQA accuracy (\%). Ovis2 and Gemma3 generally demonstrate stronger generalization, while the Quantity category proves to be a significant bottleneck for all models in this challenging zero-shot setting.}
  \label{fig:vqa_results_radar}
\end{figure*}

\subsubsection{Task Evaluation Protocol.} We evaluated two tasks: Visual Question Answering (VQA) and image captioning. For VQA, we assessed five reasoning categories: Object Recognition, Feature Analysis, Spatial Reasoning, Quantitative Queries, and Presence Verification. Due to complex, sentence-based answers, we used GPT-4.1 for automated correctness judgment, comparing the model’s response to the ground-truth answer. GPT-4.1 provided a structured judgment (<judge>Correct/Wrong</judge>) with justification. VQA accuracy is the percentage of 'Correct' judgments per category. For \textbf{image captioning}, we compared generated captions to reference captions using BBLEU-1 to B-4 (B-1 to B-4)~\cite{bleu}, ROUGE-L (R-L)~\cite{rouge}, and METEOR (M)~\cite{metor}. Inspired by LLaVA's \cite{llava} evaluation methodology, we introduced a Caption Score (CS) (0-100) from GPT-4.1, evaluating accuracy, completeness, detail, and fluency, with a structured score (<score>X</score>) and rationale.

\subsubsection{Implementation Details.} All zero-shot evaluations were conducted using publicly available open-source weights from Hugging Face for each VLM, along with their associated preprocessing pipelines where applicable. The experiments were executed on NVIDIA RTX 4090 GPUs, utilizing PyTorch as the primary deep learning framework. Specific model parameters and runtime scripts for each VLM can be found in our publicly available at \url{https://github.com/StarZi0213/RSVLM-QA/tree/main/models}.

\subsection{Results and Analysis}
\label{ssec:results_analysis}

We now present and analyze the zero-shot performance of the benchmarked VLMs on RSVLM-QA across both VQA and image captioning tasks. These findings highlight the current strengths and limitations of general-purpose VLMs when applied to the specialized field of remote sensing.

The zero-shot VQA evaluation, depicted in Figure~\ref{fig:vqa_results_radar}, reveals notable disparities in the generalization capabilities of current vision-language models (VLMs) within the remote sensing domain. Ovis2 leads as the top performer, achieving an average accuracy of 66.96\%, with strong results in Object Recognition (75.14\%), Feature Analysis (83.95\%), and Spatial Reasoning (68.98\%). Gemma3 (59.49\%) and InternVL3 (57.71\%) follow, delivering commendable yet lower overall performance. A striking challenge emerges in the \textit{Quantity} category, where even Ovis2’s leading score of 37.80\% underscores the difficulty of precise numerical and counting tasks in complex aerial scenes without prior training. This indicates a gap in robust mechanisms for fine-grained object individuation and enumeration. In contrast, \textit{Presence} verification tasks yield higher scores, such as InternVL3’s 67.06\%, likely due to the binary nature of these questions aligning with general VLM strengths. The performance of BLIP-2, averaging 29.53\%, highlights the substantial leap needed for older architectures to address this diverse and challenging benchmark. These findings benchmark state-of-the-art (SOTA) capabilities and emphasize RSVLM-QA’s value in identifying key areas for future VLM research in Earth observation, including enhanced quantitative reasoning and nuanced spatial understanding. For detailed results, refer to the Model Performance Benchmark at https://rsvlm-qa.vercel.app/.

\subsubsection{Zero-Shot Image Captioning Performance}

Table~\ref{tab:captioning_results} summarizes the zero-shot performance of the VLMs on the image captioning task, evaluated using both standard metrics and our GPT-4.1 assessed Caption Score (CS).

\begin{table}[htbp] 
  \caption{Zero-shot image captioning performance on RSVLM-QA. B-N denotes BLEU-N, R-L is ROUGE-L, M is METEOR, and Caption Score(CS) is GPT-4.1 evaluated Caption Score (0-100). Higher scores indicate better performance. Best scores per metric are highlighted in \textbf{bold}.}
  \label{tab:captioning_results} 
  \small
  \begin{tabular}{lccccc} 
    \toprule
    \textbf{Model} & \textbf{B-1} & \textbf{B-4} & \textbf{R-L} & \textbf{M} & \textbf{Caption Score} \\ 
    \midrule
    Gemma3    & 31.09 &  2.75 & 19.61 & 28.84 & 72.94 \\ 
    InternVL3 & 33.36 &  3.24 & 20.51 & 29.49 & 83.90 \\
    LLaVA     & 31.90 &  2.91 & 22.76 & 21.91 & 60.43 \\
    Qwen2.5-VL & 42.59 &  4.74 & 23.90 & 28.39 & 80.09 \\
    BLIP-2    & 13.59 &  0.77 &  8.45 &  3.65 & 25.07 \\
    Ovis2     & 42.90 & 5.17 & 24.31 & 31.65 & 87.54 \\
    \bottomrule
  \end{tabular}
\end{table}

Zero-shot captioning results, presented in Table~\ref{tab:captioning_results}, highlight varying generalization capacities of benchmarked vision-language models (VLMs). Ovis2 consistently excels, achieving an impressive Caption Score of 87.54. This reflects its robust ability to produce captions that are linguistically fluent, semantically rich, and factually aligned with complex remote sensing imagery, even without prior exposure to the domain or caption style. InternVL3 and Qwen2.5-VL also show strong performance, with Caption Scores surpassing 80. The GPT-4.1-assessed Caption Score (CS) generally aligns with traditional metrics like BLEU and METEOR, yet it offers a broader evaluation, capturing aspects such as detail and completeness beyond n-gram-based measures. BLIP-2’s notably lower performance underscores the progress of recent VLM architectures in generative tasks requiring nuanced understanding and expression. These findings reveal the challenges posed by remote sensing imagery and affirm the value of RSVLM-QA’s captioning component in assessing fine-grained descriptive capabilities.

\section{Conclusion}
\label{sec:conclusion}

In this paper, we introduce RSVLM-QA, a novel, large-scale dataset derived from multiple sources and enriched with detailed annotations to advance Remote Sensing Visual Question Answering (RSVQA). Our innovative methodology harnesses GPT-4.1 for scalable, precise annotations, adopts a dual-track VQA generation strategy to balance descriptive depth and quantitative accuracy, and implements a rigorous multi-stage curation process. Zero-shot benchmarking of six leading vision-language models (VLMs) on RSVLM-QA reveals its challenging nature and underscores its value in exposing current VLM limitations. While our LLM-based evaluation ensures scalability, we recognize its limitation compared to comprehensive human assessment. Future work will explore VLM adaptation through fine-tuning on RSVLM-QA, expand the dataset to include more diverse reasoning tasks, such as temporal dynamics, and develop novel architectures to address identified performance bottlenecks, thus driving innovation in vision-language understanding for remote sensing applications.



\end{document}